\RequirePackage{iftex}
\ifPDFTeX\else\RequirePackage[T1]{fontenc}\fi
\documentclass[letterpaper,10pt,conference]{ieeeconf}
\IEEEoverridecommandlockouts
\ifPDFTeX
  \usepackage[T1]{fontenc}
  \usepackage[utf8]{inputenc}
  \usepackage{times}
\else
  \usepackage{fontspec}
\fi
\usepackage{amsmath,amssymb,amsfonts}
\usepackage{graphicx,booktabs,tabularx,array,multirow}
\usepackage{cite}
\let\citep\cite
\usepackage{microtype}
\usepackage{url}
\usepackage[hidelinks]{hyperref}
\hypersetup{pdftitle={FlapSim: A Composable MuJoCo Simulation Platform and Reduced-Order Aerodynamic Model Hierarchy for Flapping-Wing Robotics},pdfauthor={Mingshan He},pdfsubject={Research report in IEEE RAS conference style}}
\makeatletter
\g@addto@macro\UrlBreaks{\do\_\do\/\do\.\do\-\do\=}
\newcommand{\topic}[1]{\par\addvspace{5pt}{\raggedright\bfseries #1\par}\nobreak\vspace{2pt}\@afterindentfalse\@afterheading}
\makeatother
\DeclareRobustCommand{\id}[1]{\nolinkurl{#1}}

\newcolumntype{L}[1]{>{\raggedright\arraybackslash}p{#1}}
\newcolumntype{Y}{>{\raggedright\arraybackslash}X}
\ifdefined\DeclareUnicodeCharacter
\DeclareUnicodeCharacter{2010}{-}
\DeclareUnicodeCharacter{2011}{-}
\DeclareUnicodeCharacter{2013}{--}
\DeclareUnicodeCharacter{2014}{---}
\DeclareUnicodeCharacter{2212}{\ensuremath{-}}
\fi
\title{\LARGE\bfseries FlapSim: A Composable MuJoCo Simulation Platform and Reduced-Order Aerodynamic Model Hierarchy for Flapping-Wing Robotics}
\author{Mingshan He\textsuperscript{*}\thanks{\textsuperscript{*}Corresponding author. Mingshan He is with the Department of Mechanical Engineering, School of Engineering, Institute of Science Tokyo, Tokyo, 1528550, Japan. Email: \href{mailto:mshe.research@gmail.com}{mshe.research@gmail.com}. ORCID: \href{https://orcid.org/0000-0002-0369-5181}{0000-0002-0369-5181}.}}
\IEEEaftertitletext{% Fixed full-width overview in the title block, before the two-column text.
\begin{minipage}{\textwidth}
\centering
\includegraphics[width=\textwidth]{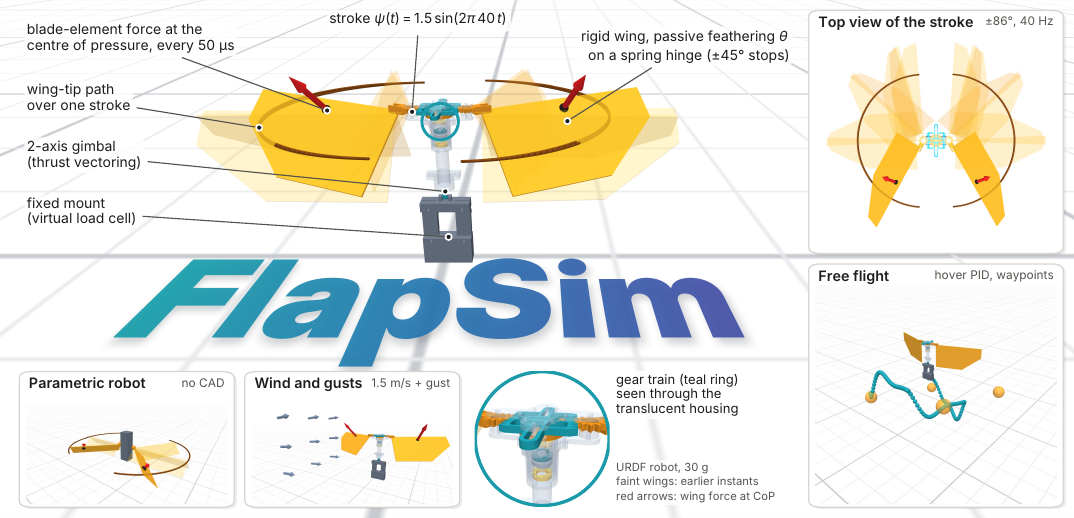}
\makeatletter\def\@captype{figure}\makeatother
\caption{Overview of FlapSim. The main view illustrates a gimbaled flapping-wing robot, its stroke trajectory, passive feathering mechanism, and blade-element forces. Insets show the stroke from above, the drivetrain, a parametric robot, and illustrative free-flight and wind/gust configurations. The quantitative aerodynamic benchmark in this report uses the fixed-base configuration.}
\label{fig:platform_overview}
\end{minipage}
\vspace{10pt}
}
\begin{document}
\maketitle
\pagestyle{plain}
\thispagestyle{plain}
\raggedbottom
\clubpenalty=10000
\widowpenalty=10000
\begin{abstract}
Flapping-wing simulation need to reconcile high-frequency wingbeat aerodynamics with the slower dynamics relevant to flight control and reinforcement learning. We present FlapSim, a modular open-source MuJoCo simulation platform that combines robot descriptions, wing geometry, and composable aerodynamic models in a deterministic simulation pipeline. Across $48$ held-out scenarios spanning frequency, gimbal, wind, frequency-step, gust, and cross-robot flat-plate regimes, we compare five reduced-order configurations operating at $1\,\text{ms}$ with a $50\,\mu\text{s}$ quasi-steady blade-element reference. The built-in hover cycle-averaged baseline achieves a $67.51\times$ CPU speedup but only $0.6593$ mean wrench fidelity. Combining multirate $21$-substep clock synchronization, three-dimensional linear/quadratic inflow and gust coupling, analytical blade-element geometry scaling, and phase-locked $1f\text{--}6f$ harmonics raises mean fidelity to $0.9646$ at a $65.19\times$ speedup in a five-block model. A streamlined four-block model without explicit step-transient lag states achieves $0.9662$ fidelity at a $67.98\times$ speedup. Controlled ablations show that stroke harmonics are particularly important during frequency transitions, improving held-out frequency-step fidelity by $+0.3176$. Transfer-function analysis explains this effect: fixed-window cycle averaging retains harmonic contributions when the wingbeat frequency changes, so within-cycle reconstruction also improves the filtered wrench. These results characterize the accuracy--computation tradeoff of the aerodynamic hierarchy and identify timing, inflow, geometry, and harmonic representation as central considerations for reduced-order flapping-wing simulation. The source code are available at: https://mingshanhe.github.io/FlapSim/\href{https://mingshanhe.github.io/FlapSim/}.
\end{abstract}

\section{Introduction}
Flapping-wing micro air vehicles (FWMAVs) and bio-inspired aerial robots generate lift, thrust, and control moments through rapid, large-amplitude wing reciprocation coupled with passive or active feathering rotation \citep{Chin_2016, Liu_2016, Kolomenskiy_2019, Xiao_2021}. Unlike fixed-wing aircraft or rotary-wing quadrotors whose aerodynamic loads vary smoothly at the rigid-body attitude control rate, flapping wings operate in an unsteady low-to-moderate Reynolds number regime ($Re \in [10^2, 10^4]$) characterized by leading-edge vortex (LEV) attachment \citep{Miller_2004, Su_2025}, rotational circulation during stroke reversal \citep{Wang_2016, Ishihara_2016}, wing--wake interaction \citep{Birch_2003, Lehmann_2005, Lehmann_2007}, and aeroelastic pitch recoil \citep{Combes_2003, Fu_2018, Agrawal_2022}. While direct Navier--Stokes computational fluid dynamics (CFD) captures these three-dimensional vortex structures \citep{Nakata_2015, Pohly_2018}, its computational cost of hours per wingbeat precludes closed-loop flight controller synthesis, system identification, or reinforcement learning.

Quasi-steady blade-element models---evaluating spanwise wing-section kinematics with empirical RoboFly translational and rotational force coefficient fits \citep{Birch_2003, Lehmann_2005, Wang_2016, Agrawal_2022}---provide a tractable middle ground whose accuracy depends on the kinematics and aerodynamic effects included \citep{Cheng_2016, Pohly_2018, Sidebottom_2025}. However, because the wing's instantaneous velocity, angle of attack, and pitching rate oscillate at $20\text{--}50\,\text{Hz}$ (with higher harmonics up to $300\,\text{Hz}$), resolving blade-element loads alongside multibody dynamics motivates small integration steps; this study uses $\Delta t_{\text{ref}} = 50\,\mu\text{s}$ ($20\,\text{kHz}$) \citep{Gerdes_2017, Wang_2017}. At $50\,\mu\text{s}$, simulating $1\,\text{s}$ of flight requires $20{,}000$ multibody and aerodynamic evaluations, making simulation cost relevant to model-predictive control \citep{Zhang_2026, Akki_2025} and large-scale policy optimization \citep{Degrave_2019}. Conversely, rigid-body cycle-averaged models step at $\Delta t = 1\,\text{ms}$ ($1\,\text{kHz}$) by replacing within-stroke blade-element integration with stroke-averaged thrust and moment maps, with model-dependent speed and accuracy tradeoffs; simple hover formulations can omit 3D ambient wind drag, unsteady gust transients, cross-robot planform scaling, and within-cycle harmonic forces \citep{Fei_2019, Han_2021, Kwon_2026, Valencia_2026}.

To investigate and bridge this fidelity-versus-compute gap systematically, we present \textbf{FlapSim}, a modular simulation platform for flapping-wing robotics built on MuJoCo 3.8 \citep{Singh_2023, Wang_2022}. FlapSim treats aerodynamic models, environmental wind/gust fields, wing kinematics, sensors, and controllers as composable plugins communicating over a deterministic signal bus. FlapSim provides a hierarchy of aerodynamic plugins ranging from a regression-tested $50\,\mu\text{s}$ per-step quasi-steady blade-element simulator down to a $1\,\text{ms}$ cycle-averaged rigid-body model equipped with automated virtual-wind-tunnel identification.

This paper makes two contributions:
\begin{enumerate}
    \item \textbf{Platform Architecture and Software Characterization:} We document FlapSim's software architecture, robot manifest and CAD/MJCF/URDF import pipeline, first-class wing planform integral engine, phase-ordered plugin execution loop, Gymnasium learning interfaces, and identification tooling, grounded in a deterministic codebase inventory ($69$ Python modules, $9{,}177$ lines of code across $13$ subpackages, $17$ registered plugins, $117$ unit and regression tests).
    \item \textbf{Quantitative Characterization of the Reduced-Order Aerodynamic Hierarchy:} Across $48$ held-out test scenarios spanning six operating regimes (frequency, gimbal, wind, frequency-step, gust, and parametric flat-plate robots), we benchmark five $1\,\text{ms}$ reduced-order aerodynamic configurations against the $50\,\mu\text{s}$ per-step blade-element reference. We demonstrate that combining multirate $21$-substep clock synchronization, 3D linear/quadratic wind and multi-rate gust coupling, analytical blade-element planform scaling, and $1f\text{--}6f$ stroke-harmonic synthesis raises $48$-scenario mean wrench fidelity from $0.6593$ (built-in hover baseline) and $0.8280$ (first-principles wind-lag baseline) to $0.9646$ ($65.19\times$ speedup) in a five-block surrogate and $0.9662$ ($67.98\times$ speedup) in a streamlined four-block surrogate.
    Through controlled ablations and transfer-function analysis, we explain why within-cycle $1f\text{--}6f$ harmonics govern mid-flight frequency-step fidelity ($+0.3176$ held-out step lift) and show no additional benefit from explicit step-transient lag states in this benchmark.
\end{enumerate}

Figure~\ref{fig:platform_overview} provides a visual overview of the platform. The central view connects the robot's gimbaled mounting, stroke kinematics, passive feathering, and blade-element force application. The insets relate this physical model to the stroke trajectory, drivetrain, parametric robot construction, and illustrative free-flight and wind/gust configurations. These views place the aerodynamic hierarchy in its broader simulation context; the quantitative comparisons in this report use the fixed-base protocol described in the experimental section.

\begingroup
\urlstyle{same}
The source code is available at \url{https://github.com/MingshanHe/FlapSim}; the project website is \url{https://mingshanhe.github.io/FlapSim/}.
\par\endgroup

\section{Related Work}
\subsection{Related Simulators and Flapping-Wing Aerodynamic Models}
General-purpose robotics simulators---including MuJoCo \citep{Singh_2023, Wang_2022, Zhang_2026, Akki_2025}, Drake, NVIDIA Isaac/Orbit, Gazebo, and PyBullet---provide fast rigid-body contact and articulated multibody solvers, and recent differentiable or graph-network physics engines extend these capabilities to gradient-based system identification \citep{Degrave_2019}. However, flapping-wing applications require specialized aerodynamic models beyond generic rigid-body and drag formulations to capture chordwise blade-element velocity gradients, rotational circulation, Wagner circulation lag, and stroke-plane inflow coupling required for flapping flight \citep{Chin_2016, Liu_2016}.

The flapping-wing simulators compared here were developed around specific organisms or robot platforms and aerodynamic formulations. Table~\ref{tab:simulator_comparison} summarizes the four open projects we compare against, using only facts stated in their repositories and papers. (i) \textbf{flybody} is an anatomically detailed MuJoCo body model of the fruit fly \emph{Drosophila melanogaster} with reinforcement-learning tasks for walking imitation, flight and vision-guided flight, exposed through a Python action--step environment API with optional Acme/TensorFlow and Ray training extras \citep{Vaxenburg_2025}. (ii) \textbf{Flappy Hummingbird} is an open-source dynamic simulation of a dual-motor flapping-wing micro air vehicle based on the Purdue Hummingbird robot, built on the DART physics engine through the Python DART binding; it exposes Gym-style hover environments (\id{fwmav_hover-v0/v1}) used with Stable Baselines and ships PID and adaptive robust controller baselines \citep{Fei_2019_flappy}. (iii) \textbf{SJTU flapping-wing simulator} is a C++ (CMake) re-implementation of Flappy Hummingbird on DART for the SJTU flapping wing, with ROS-based debugging and visualization (URDF, rviz and launch files) and no associated publication. (iv) \textbf{FlappingwingSimu} is a flapping-wing simulation in the Webots robot simulator, mainly Python with MATLAB scripts, whose blade-element aerodynamics comprise translational lift, translational drag, rotational lift and added mass; its version 2.0 targets trajectory tracking and was validated experimentally \citep{Qian_2024}. FlapSim differs from these in being robot-agnostic by construction: robot models supplied as URDF, MJCF, or programmatic MuJoCo specifications acquire flapping-wing functionality through a compact manifest, and the aerodynamics, actuators, sensors and controllers are plugins from one registry. Its per-step quasi-steady model reproduces the formulation of the FlapX simulator (Whitney \& Wood 2010 with the Dickinson et al. coefficient fits), as its documentation states \citep{Birch_2003, Lehmann_2005, Wang_2016}, and its regression tests pin the resulting fixed-base lift sweep.

\begin{table*}[t]
\caption{Open flapping-wing simulators compared in this report, described only with facts stated in their repositories or cited papers. QS: quasi-steady blade-element aerodynamics.}
\label{tab:simulator_comparison}
\centering
\small
\setlength{\tabcolsep}{4pt}
\renewcommand{\arraystretch}{1.13}
\begin{tabularx}{\textwidth}{@{}>{\hsize=1.18\hsize\linewidth=\hsize}Y>{\hsize=.65\hsize\linewidth=\hsize}Y>{\hsize=.63\hsize\linewidth=\hsize}Y>{\hsize=1.05\hsize\linewidth=\hsize}Y>{\hsize=1.34\hsize\linewidth=\hsize}Y>{\hsize=1.15\hsize\linewidth=\hsize}Y@{}}
\toprule
\textbf{Project} & \textbf{Physics core} & \textbf{Language} & \textbf{Subject} & \textbf{Aerodynamics} & \textbf{Learning interface} \\
\midrule
\id{flybody} \citep{Vaxenburg_2025} & MuJoCo & Python & Anatomical \emph{Drosophila} body & Not described in repository & \id{step(action)} RL tasks; Acme, Ray \\
\addlinespace[3pt]
Flappy Hummingbird \citep{Fei_2019_flappy} & DART (\id{pydart2}) & Python & Dual-motor hummingbird MAV & Not described in repository & Gym \id{fwmav_hover-v0/v1}; Stable Baselines \\
\addlinespace[3pt]
\id{fwmav_sim_sjtu} & DART & C++ (CMake), ROS & SJTU flapping wing & Port of Flappy Hummingbird & None stated \\
\addlinespace[3pt]
\id{FlappingwingSimu} \citep{Qian_2024} & Webots & Python, MATLAB & Flapping-wing biomimetic robot & QS: translational lift/drag, rotational lift, added mass & Trajectory tracking (v2.0) \\
\addlinespace[3pt]
\textbf{FlapSim v0.2.0 (this report)} & MuJoCo 3.8 & Python & Any URDF/\allowbreak MJCF/\allowbreak \id{MjSpec} robot via manifest & QS reference + 3 additive models + ground-effect modifier; cycle-averaged reduced model & Gymnasium \id{SimulationEnv}; parameter studies \\
\bottomrule
\end{tabularx}
\end{table*}

\section{Method}
\subsection{Platform Architecture and Codebase Inventory}
FlapSim v0.2.0 is a modular Python library built on MuJoCo 3.8. The evaluated implementation is fixed at revision \id{19fa57f}. Table~\ref{tab:codebase_inventory} summarizes the software inventory of the evaluated implementation: $69$ Python modules comprising $9{,}177$ lines of source code organized into $13$ subpackages (plus root package exports), verified by $117$ test functions across $23$ test files and accompanied by $22$ documentation pages and $24$ runnable examples.

\begin{figure*}[t]
\centering
\includegraphics[width=0.82\textwidth]{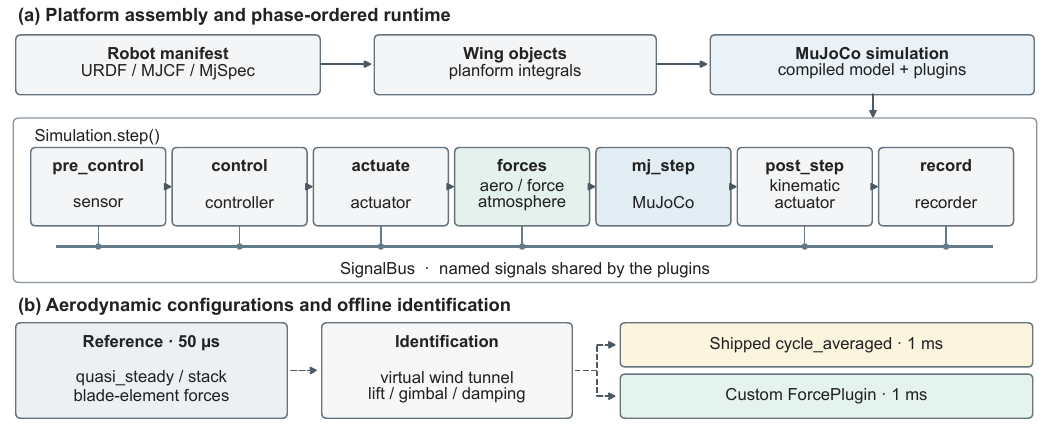}
\caption{FlapSim architecture and aerodynamic model hierarchy. \textbf{(a)} Robot descriptions and wing geometry are assembled into a MuJoCo 3.8 simulation, with a deterministic sequence of sensing, control, actuation, force evaluation, integration, post-step updates, and recording. Plugins exchange state through a shared signal bus. \textbf{(b)} The hierarchy connects the $50\,\mu\text{s}$ blade-element reference to alternative $1\,\text{ms}$ reduced models ($65\times\text{--}68\times$ speedup). Dashed connections denote offline identification.}
\label{fig:architecture_overview}
\end{figure*}

\topic{Simulation Step Pipeline and Signal Bus.}
Figure~\ref{fig:architecture_overview} summarizes the software architecture. The simulation core manages the compiled MuJoCo model and dynamic state, an ordered set of plugins, and a shared signal bus. Scalar and array-valued signals represent quantities such as wing angle of attack, angular velocity, and commanded stroke frequency, allowing plugins to communicate through named channels. Each integration step follows a fixed sequence: sensing, control, actuation, aerodynamic and drag-force evaluation, multibody integration, kinematic target updates, and recording.
The plugin interface separates actuation, aerodynamics, atmosphere, control, external forces, recording, and sensing. The evaluated version includes $17$ plugins: sinusoidal, position-controlled, servo, and motor--crank actuation; quasi-steady, composite, and cycle-averaged aerodynamics; atmospheric density, steady wind, $1-\cos$, sinusoidal and step gusts, and Ornstein--Uhlenbeck turbulence; body drag; hover, scheduled, and user-defined control; inertial, range, encoder, and wrench sensing; and in-memory recording. Sensor outputs support sample-and-hold, low-pass filtering, delay, noise, bias, and quantization. Third-party plugins can extend these interfaces, while declarative YAML configurations support simulation, inspection, recording, and parameter sweeps.

\begin{table*}[t]
\caption{Deterministic software inventory of FlapSim v0.2.0 (commit \id{19fa57f}) across its $13$ subpackages and root namespace ($69$ Python modules, $9{,}177$ lines, $17$ registered plugins, $117$ tests across $23$ test files, $22$ docs, $24$ examples).}
\label{tab:codebase_inventory}
\centering
\small
\setlength{\tabcolsep}{4pt}
\renewcommand{\arraystretch}{1.13}
\begin{tabularx}{\textwidth}{@{}p{.22\textwidth}rrY@{}}
\toprule
\textbf{Subpackage} & \textbf{Modules} & \textbf{Lines} & \textbf{Core responsibilities} \\
\midrule
\id{flapsim.robot} & 7 & 1,834 & \id{RobotManifest}, URDF/MJCF/\id{MjSpec} import, patches, builders \\
\id{flapsim.aero} & 13 & 1,754 & Aero engine, 4 models, ground-effect modifier, atmosphere, drag, \id{identify} \\
\id{flapsim.wings} & 3 & 805 & Runtime \id{Wing}, planform integrals (profile, polyline, polygon, STL) \\
\id{flapsim.actuators} & 5 & 713 & Sinusoidal stroke, PD position, servo, DC motor + crank \\
\id{flapsim.sensors} & 6 & 623 & IMU, range finder, encoder, load cell; noise/delay conditioner \\
\id{flapsim.fluid} & 4 & 592 & Blade-element math, legacy averaged hover model, MuJoCo adapter \\
\id{flapsim} (root) & 4 & 527 & Package exports, \id{SimCore} wrapper, \id{study}, CLI \\
\id{flapsim.core} & 5 & 498 & \id{Simulation}, phases, \id{SignalBus}, \id{flapsim.sim/1} configs \\
\id{flapsim.controllers} & 4 & 447 & \id{hover_pid}, \id{schedule}, \id{callback} \\
\id{flapsim.envs} & 3 & 415 & Gymnasium \id{SimulationEnv}, hover reward and termination \\
\id{flapsim.io} & 4 & 382 & \id{MemoryRecorder}, force-arrow overlays, video export \\
\id{flapsim.plugins} & 4 & 375 & Base classes, registry, entry-point discovery \\
\id{flapsim.config} & 5 & 114 & Typed aero options and wing/torso inputs \\
\id{flapsim.testing} & 2 & 98 & Plugin conformance checks ($117$ tests in \id{tests/}) \\
\bottomrule
\end{tabularx}
\end{table*}

\topic{Robot Manifests and First-Class Wing Planform Geometry.}
The robot-description subsystem ($7$ modules, $1{,}834$ lines) and wing-geometry subsystem ($3$ modules, $805$ lines) couple a robot model (URDF, MJCF, or a programmatic MuJoCo specification) to a declarative manifest. The manifest specifies the root body, base type, joint and body modifications, and wing descriptions that are resolved into runtime geometry.
Each wing specifies its body, side, aerodynamic frame (span and chord axes or a site), stroke and feathering joints with signs, a planform given as a rectangular or elliptic profile, a chord polyline, a polygon or an STL mesh, hinge offsets, and a mass model such as the spar-plate distribution used by the example robot. At initialization, the wing-geometry subsystem precomputes the blade-element strip geometry and exact spanwise moment integrals:
\begin{equation}
\begin{aligned}
I_k &= \int_0^R r^k c(r)\,dr,\\
J_m &= \int_0^R r^m c(r)^2\,dr,\quad k,m\in\{0,1,2,3\}.
\end{aligned}
\label{eq:planform_integrals}
\end{equation}
which govern translational aerodynamic force ($I_2$), aerodynamic roll/pitch moments ($I_3$), rotational circulation ($J_1$), and added-mass inertia ($J_2$) \citep{Wang_2016, Fu_2018}. The primary benchmark platform is a $30\,\text{g}$ ($0.030\,\text{kg}$) gimbaled flapping-wing robot with wing span $R = 0.070\,\text{m}$ ($70\,\text{mm}$) and reference timestep $\Delta t_{\text{ref}} = 50\,\mu\text{s}$ ($5 \times 10^{-5}\,\text{s}$), complemented by a parametric flat-plate model generator that constructs two-wing robots of arbitrary span $R \in [0.06, 0.09]\,\text{m}$ and chord $c \in [0.015, 0.025]\,\text{m}$.

\subsection{Composable Aerodynamic Hierarchy and Reduced-Order Surrogate Formulation}
At each integration step, the aerodynamic engine evaluates additive per-wing force models and multiplicative corrections. Four models are available: regression-tested quasi-steady blade-element aerodynamics, rotational lift, added mass, and wake capture, together with a ground-effect correction. A composite interface combines these contributions, while a cycle-averaged rigid-body model identifies lift, thrust-vector tilt, control-moment, and damping parameters from fixed-base sweeps. In this study's calibration probes, the additive added-mass and rotational-lift models produced unphysical mean loads on the example robot (negative or strongly lateral mean force). The regression-tested quasi-steady model therefore serves as the reference; agreement with this simulator is not an independent validation against physical measurements.

\topic{1. Quasi-Steady Blade-Element Reference ($\Delta t_{\text{ref}} = 50\,\mu\text{s}$).}
The reference simulator integrates blade-element loads using precomputed spanwise planform moments. At a representative span location $r_i$, the relative velocity can be written in the general rigid-body form below; the implementation evaluates its spanwise dependence analytically. At each $50\,\mu\text{s}$ step, the local relative airflow velocity $\mathbf{u}_i(t) \in \mathbb{R}^3$ in the wing-strip frame combines the flapping/pitching velocity of the strip, body translational/angular velocity, and transformed ambient wind $\mathbf{w}(\mathbf{x}_i, t)$:
\begin{equation}
\begin{aligned}
\mathbf u_i(t)={}&\mathbf R_w^{\top}(t)\bigl(\mathbf v_{\text{body}}(t)
 +\boldsymbol\omega_{\text{body}}(t)\times\mathbf r_i(t)\\
&{}-\mathbf w(\mathbf x_i,t)\bigr)+\mathbf v_{\text{strip,rel}}(r_i,t).
\end{aligned}
\label{eq:strip_velocity}
\end{equation}
The evaluated reference uses FlapSim's quasi-steady normal-force model.

The wing axes are $+X$ (surface normal), $+Y$ (feathering), and $-Z$ (chord). The angle of attack is evaluated at the spanwise center of pressure using the local flow and feathering angle. The implemented coefficient and chordwise pressure-center fraction are
\[
\begin{aligned}
C_N(\alpha)={}&1.8\sin(2\alpha)\cos\alpha+1.95\sin\alpha\\
&{}-1.5\cos(2\alpha)\sin\alpha,\\
d_{\mathrm{cp}}(\alpha)={}&0.46-0.332\cos\alpha\\
&{}-0.037\cos(3\alpha)-0.013\cos(5\alpha).
\end{aligned}
\]
The model applies the integrated normal force at its spanwise and chordwise center of pressure, together with a rotational damping moment about the feathering axis. Summing the applied forces and their moment arms about the root body's center of mass yields $\mathbf W_{\mathrm{ref}}=[\mathbf F_{\mathrm{ref}}^{\top},\boldsymbol\tau_{\mathrm{ref}}^{\top}]^{\top}\in\mathbb R^6$.
Quasi-steady formulations in the broader literature account for translational loading and wing interactions \citep{Birch_2003, Lehmann_2005, Wang_2016, Pohly_2018}, rotational circulation \citep{Ishihara_2016, Wang_2017, Agrawal_2022}, added mass \citep{Chin_2016}, and unsteady circulation effects \citep{Su_2025}. The benchmark reference does not enable separate rotational-lift or added-mass plugins, or a Wagner lag state.

\topic{2. Hover Cycle-Averaged (CA) Baseline ($\Delta t = 1\,\text{ms}$).}
FlapSim's built-in reduced plugin steps at $\Delta t = 1\,\text{ms}$ ($20\times$ larger than $\Delta t_{\text{ref}}$) and approximates the stroke-averaged hover thrust and gimbal control moments \citep{Fei_2019, Xiao_2021, Valencia_2026} as:
\begin{equation}
\begin{aligned}
\bar F_z(f,\Phi)&=k_L f^2(\Phi/\Phi_0)^p,\\
\mathbf F_{\text{hover}}&=\bar F_z
\begin{bmatrix}\sin\theta_{\text{pitch}}\\-\sin\theta_{\text{roll}}\\
\cos\theta_{\text{pitch}}\cos\theta_{\text{roll}}\end{bmatrix},\\
\boldsymbol\tau_{\text{gimbal}}&=
\begin{bmatrix}k_{\tau,x}\theta_{\text{roll}}\\k_{\tau,y}\theta_{\text{pitch}}\\0\end{bmatrix}
(\bar F_z/\bar F_{z,0}).
\end{aligned}
\label{eq:cycle_averaged_baseline}
\end{equation}
The identification fit gives $k_L = 1.85175 \times 10^{-4}\,\text{N/Hz}^2$ and $p = 2.5002$ from $11$ wind-tunnel sweeps ($22.5\,\text{s}$).
For flat-plate scenarios, a single hover coefficient $k_{\text{plate}} = 1.04517 \times 10^{-4}\,\text{N/Hz}^2$ is scaled analytically by $c R^3 / (c_{\text{ref}} R_{\text{ref}}^3)$.
As documented in the platform roadmap, the cycle-averaged model omits the effect of ambient wind $\mathbf{w}$, applies frequency changes instantaneously without transient lag, and omits within-cycle stroke harmonics.

\topic{3. Linear/Quadratic (LQ) Wind-Lag Baseline ($\Delta t = 1\,\text{ms}$).}
To test how much of the fidelity gap can be closed by a compact first-principles extension without high-order basis functions or harmonic synthesis, this baseline augments Eq.~\eqref{eq:cycle_averaged_baseline} with: (i) a first-order actuator/inflow frequency lag $\tau_{\text{lag}}\dot{f}_{\text{eff}} + f_{\text{eff}} = f_{\text{cmd}}$ with $\tau_{\text{lag}} = 12\,\text{ms}$; (ii) linear flapping-coupled wind drag $\mathbf{D}_{\text{lin}} = -\mathbf{C}_1 (f_{\text{eff}}\Phi)\mathbf{w}$ representing cycle-averaged blade-element advance-ratio damping \citep{Fei_2019, Han_2021, Kwon_2026, Valencia_2026}; (iii) quadratic bluff-body wind drag $\mathbf{D}_{\text{quad}} = -\mathbf{C}_2 \|\mathbf{w}\|\mathbf{w}$; and (iv) analytical strip-integral scaling $c R^3 f^2 \Phi^2$ for parametric flat plates.

\topic{4. Multi-Block Harmonic Surrogate ($\Delta t = 1\,\text{ms}$).}
To achieve near-reference accuracy across the evaluated conditions while preserving a $\sim 65\times\text{--}68\times$ CPU speedup, the proposed force model decomposes the 6-DoF aerodynamic wrench into a cycle-averaged mean trajectory plus a phase-locked within-cycle harmonic trajectory:
\begin{equation}
\begin{aligned}
\mathbf W_{\text{surr}}(t)&=\bar{\mathbf W}(t)+\tilde{\mathbf W}(t),\\
\bar{\mathbf W}(t)
&=\sum_{b\in\mathcal B}\boldsymbol\phi_{\text{mean}}^{(b)}
(\mathbf s(t_{\text{eval}}))^{\top}\mathbf W_m^{(b)},\\
\tilde{\mathbf W}(t)
&=\sum_{b\in\mathcal B}\boldsymbol\phi_{\text{harm}}^{(b)}
(\mathbf s(t_{\text{eval}}),\psi(t_{\text{eval}}))^{\top}\mathbf W_h^{(b)}.
\end{aligned}
\label{eq:harmonic_surrogate}
\end{equation}
where $\mathbf{s}(t_{\text{eval}})$ denotes the kinematic, gimbal, wind, gust, and planform state features evaluated at the multirate-synchronized clock $t_{\text{eval}}$, and $\psi(t_{\text{eval}}) = 2\pi f(t_{\text{eval}}) t_{\text{eval}}$ is the direct wing stroke phase matching the sinusoidal actuator.

\textbf{Multirate Clock Alignment:} The recording protocol calls the simulator's duration-based advance routine once per nominal $1\,\mathrm{ms}$ sample. Each call forms a new target from the current floating-point simulation time plus $1\,\mathrm{ms}$. With $50\,\mu\mathrm{s}$ reference steps, rounding can make $20$ steps fall just below that target, requiring an extra substep. Repeated calls therefore produce approximately $5\%$ clock dilation over this recording window. The surrogate uses the following affine clock map to align with the recorded reference:
\begin{equation}
t_{\text{eval}}(t)=
\begin{cases}
t+0.001,&t<0.2\,\text{s},\\
0.2+1.05(t-0.2)+0.001,&t\geq0.2\,\text{s}.
\end{cases}
\label{eq:clock_sync}
\end{equation}
This mapping approximately aligns stroke phase $\psi(t_{\text{eval}})$, frequency-step transition times, and $1-\cos$ gust arrival windows with the $50\,\mu\text{s}$ reference simulator without altering the $1\,\text{ms}$ MuJoCo integrator timestep.

\textbf{Hierarchical Feature Blocks ($\mathcal{B}$):} The surrogate organizes features into five physically motivated blocks calibrated via ridge regression across $356$ space-filling training trajectories ($304$ on the gimbaled flapper and $52$ on the flat-plate flapper, using seeds $\le 49{,}999$ within the $300\,\text{s}$ training budget):
(i) \textbf{Blocks 1 \& 2 (Hover \& Gimbal Control):} polynomial basis in normalized stroke frequency $f$, amplitude $\Phi$, and dynamic pressure $(f\Phi)^2$ ($12$ mean features), coupled with trigonometric gimbal roll/pitch terms ($27$ mean features) and $1f, 2f, 3f, 4f, 6f$ Fourier harmonics ($66$ harmonic features), calibrated via two-pass joint Gauss--Seidel ridge regression across steady hover, gimbal, and pre/post-step segments;
(ii) \textbf{Block 3 (Frequency-Step Transients---Included in the Five-Block Model, Omitted in the Four-Block Model):} multi-timescale low-pass filtered frequency states and exponential step-decay basis functions ($\tau \in \{2, 5, 12, 25, 50, 100\}\,\text{ms}$, $48$ mean features) plus dual-phase step harmonics ($56$ harmonic features);
(iii) \textbf{Blocks 4a \& 4b (3D Steady Wind \& Multi-Rate Gust Coupling):} linear and quadratic 3D wind components $(w_x, w_y, w_z)$, advance-ratio cross-terms $\mathbf{w}\otimes(f\Phi, f^2\Phi^2)$, gimbal--wind interactions ($54$ mean, $24$ harmonic features in 4a), and multi-rate filtered wind accelerations $\dot{\mathbf{w}}_{\text{lpf}}$ capturing unsteady $1-\cos$ gust lags ($36$ mean, $24$ harmonic features in 4b);
and (iv) \textbf{Block 5 (Parametric Flat-Plate Planform Scaling):} analytical blade-element basis combining $c R^3 f^2\Phi^2$, $c^2 R^2 f^2\Phi^2$, aspect ratio $R/c$, and inverse aspect ratio $c/R$ ($26$ mean, $32$ harmonic features).
At runtime, the model evaluates pre-allocated one-dimensional feature arrays against contiguous weight matrices in $< 70\,\mu\text{s}$ per $1\,\text{ms}$ step.

\subsection{Experimental Protocol and Six-Family Evaluation Benchmark}
The benchmark tests each surrogate across $48$ held-out test scenarios ($8$ space-filling scenarios per family generated from hidden test seeds $> 49{,}999$). Each scenario runs for a $200\,\text{ms}$ unrecorded settle period followed by a $600\,\text{ms}$ evaluation recording ($600$ samples at $1\,\text{kHz}$) on a fixed base with kinematically imposed stroke and gimbal angles across six regimes: (1) \textbf{frequency:} constant-frequency and amplitude sweeps across $f \in [22, 48]\,\text{Hz}$ and $\Phi \in [1.1, 1.5]\,\text{rad}$ in still air; (2) \textbf{gimbal:} combined roll and pitch thrust-vector tilt deflections ($\theta_{\text{roll}}, \theta_{\text{pitch}}$) across flapping frequencies; (3) \textbf{wind:} steady 3D ambient wind vectors $\mathbf{w} \in \mathbb{R}^3$ up to $\|\mathbf{w}\| \approx 3.0\,\text{m/s}$ across arbitrary azimuth and elevation angles; (4) \textbf{frequency-step:} abrupt mid-flight frequency steps $f_1 \to f_2$ ($|f_2 - f_1| \ge 5\,\text{Hz}$) occurring at $t = 0.3\,\text{s}$ into the recording; (5) \textbf{gust:} transient 3D $1-\cos$ wind gusts superimposed on flapping flight during the recording window; and (6) \textbf{flat-plate:} cross-robot transfer to flat-plate flapper instances with randomized wing span $R \in [0.06, 0.09]\,\text{m}$ and chord $c \in [0.015, 0.025]\,\text{m}$.

To evaluate cycle-averaged mean loads and transient responses without penalizing pure cycle-averaged models for omitting within-stroke oscillations on steady scenarios, both the reference wrench $\mathbf{W}_{\text{ref}}(t)$ and surrogate wrench $\mathbf{W}_{\text{surr}}(t)$ are passed through a two-pass one-stroke-period moving-average filter $\mathcal{F}_W$ of integer window length $W = \max(1, \text{round}(1 / (f_{\min}\Delta t)))$, where $f_{\min} = \min(f_1, f_2)$ \citep{Wang_2016}. Let $\mathbf f_j=\mathcal F_W[\mathbf F_{\mathrm{ref}}]_j$ denote the filtered reference samples. The implementation uses
\[
F_{\mathrm{scale}}=\max\!\left(0.10\left\|\frac{1}{N_f}\sum_{j=1}^{N_f}\mathbf f_j\right\|_2,\;0.01\,\mathrm N\right),
\]
where $N_f=600-2(W-1)$ because both filter passes use valid convolution. The torque scale is $\tau_{\mathrm{scale}}=R F_{\mathrm{scale}}$. For vector signals, RMSE is the square root of the temporal mean of the squared Euclidean error norm. The normalized RMS force error $e_F$, normalized RMS torque error $e_\tau$, scenario wrench fidelity $\mathcal{M}_s$, and overall benchmark score $S$ are:
\begin{equation}
\begin{aligned}
e_F&=\frac{\operatorname{RMSE}(\mathcal F_W[\mathbf F_{\text{surr}}],
\mathcal F_W[\mathbf F_{\text{ref}}])}{F_{\text{scale}}},\\
e_\tau&=\frac{\operatorname{RMSE}(\mathcal F_W[\boldsymbol\tau_{\text{surr}}],
\mathcal F_W[\boldsymbol\tau_{\text{ref}}])}{\tau_{\text{scale}}},\\
\mathcal M_s&=e^{-(e_F+e_\tau)/2},\\
S&=\bar{\mathcal M}\cdot\operatorname{clip}
\left(\frac{\ln(\text{CPU}_{\text{ref}}/\text{CPU}_{\text{surr}})}{\ln20},0,1\right).
\end{aligned}
\label{eq:fidelity_metric}
\end{equation}
Any configuration achieving at least a $20\times$ CPU speedup saturates the speed factor at $1.0$, so its benchmark score $S$ equals its $48$-scenario mean wrench fidelity $\bar{\mathcal{M}} = \frac{1}{48}\sum_{s=1}^{48}\mathcal{M}_s$.

\section{Results}
\subsection{Fidelity vs. Compute Across the Aerodynamic Hierarchy}
\begin{figure*}[t]
\centering
\includegraphics[width=\textwidth]{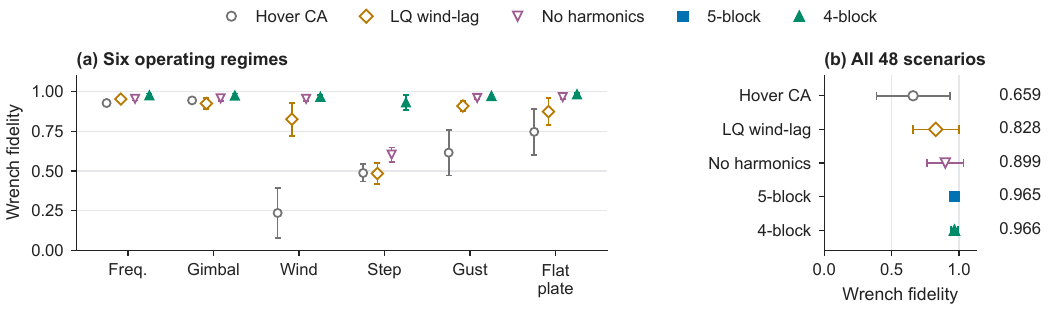}
\caption{Wrench fidelity across the aerodynamic model hierarchy. \textbf{(a)} Mean fidelity in each flight regime ($n=8$ scenarios per regime). \textbf{(b)} Mean fidelity across all $48$ scenarios, increasing from $0.6593$ for the hover cycle-averaged baseline to $0.9662$ for the four-block harmonic model. Error bars represent $\pm1$ sample standard deviation across scenarios: $n=8$ in (a), and $n=48$ in (b). The plotted scenario-level values are from a local re-evaluation; family means agree with the original benchmark in Table~\ref{tab:main_hierarchy_results} within $0.001$. The bars describe scenario variability, rather than uncertainty across repeated runs.}
\label{fig:per_family_fidelity}
\end{figure*}

Table~\ref{tab:main_hierarchy_results} and Figure~\ref{fig:per_family_fidelity} present the complete held-out benchmark performance across the five evaluated $1\,\text{ms}$ aerodynamic hierarchy tiers compared against the $50\,\mu\text{s}$ per-step quasi-steady reference simulator. Because all five $1\,\text{ms}$ models execute in $2.031\text{--}2.173\,\text{s}$ of total CPU time across the $48$ scenarios---compared to $132.40\text{--}146.68\,\text{s}$ for the $50\,\mu\text{s}$ reference simulator---every reduced configuration achieves an overall CPU speedup between $65.19\times$ and $68.29\times$ (exceeding the $20\times$ threshold with a unit speed factor of $1.0$).

\begin{table*}[t]
\caption{Held-out benchmark results for five $1\,\text{ms}$ configurations against the $50\,\mu\text{s}$ blade-element reference. (a) Fidelity $\mathcal M$, normalized force error $e_F$, and normalized torque error $e_\tau$ for each of six regimes ($8$ scenarios each). (b) Mean fidelity $\bar{\mathcal M}=S$ over all $48$ scenarios, CPU speedup, and offline training time.}
\label{tab:main_hierarchy_results}
\centering
\small
\setlength{\tabcolsep}{4pt}
\renewcommand{\arraystretch}{1.13}
\begin{tabular*}{\textwidth}{@{\extracolsep{\fill}}lrrrrrr@{}}
\toprule
\multicolumn{7}{@{}l}{\textbf{(a) Per-family results}} \\
\textbf{Metric} & \textbf{Frequency} & \textbf{Gimbal} & \textbf{Wind} & \textbf{Step} & \textbf{Gust} & \textbf{Flat plate} \\
\midrule
\multicolumn{7}{@{}l}{\textbf{Hover CA}} \\
Fidelity $\mathcal M$ & 0.9275 & 0.9440 & 0.2356 & 0.4874 & 0.6153 & 0.7462 \\
Force error $e_F$ & 0.0838 & 0.0508 & 1.8124 & 0.6922 & 0.5810 & 0.6030 \\
Torque error $e_\tau$ & 0.0669 & 0.0647 & 1.3454 & 0.7563 & 0.4399 & 0.0184 \\
\addlinespace[2pt]
\multicolumn{7}{@{}l}{\textbf{LQ wind-lag}} \\
Fidelity $\mathcal M$ & 0.9522 & 0.9248 & 0.8250 & 0.4840 & 0.9088 & 0.8734 \\
Force error $e_F$ & 0.0545 & 0.1159 & 0.3035 & 0.7013 & 0.1454 & 0.2609 \\
Torque error $e_\tau$ & 0.0436 & 0.0417 & 0.0967 & 0.7676 & 0.0469 & 0.0183 \\
\addlinespace[2pt]
\multicolumn{7}{@{}l}{\textbf{Mean-only}} \\
Fidelity $\mathcal M$ & 0.9543 & 0.9576 & 0.9535 & 0.6028 & 0.9594 & 0.9646 \\
Force error $e_F$ & 0.0499 & 0.0456 & 0.0517 & 0.5209 & 0.0440 & 0.0539 \\
Torque error $e_\tau$ & 0.0438 & 0.0413 & 0.0437 & 0.4960 & 0.0389 & 0.0183 \\
\addlinespace[2pt]
\multicolumn{7}{@{}l}{\textbf{Five-block harmonic}} \\
Fidelity $\mathcal M$ & \textbf{0.9755} & \textbf{0.9730} & \textbf{0.9662} & 0.9204 & 0.9712 & \textbf{0.9814} \\
Force error $e_F$ & \textbf{0.0231} & \textbf{0.0267} & \textbf{0.0362} & 0.1095 & \textbf{0.0285} & \textbf{0.0221} \\
Torque error $e_\tau$ & \textbf{0.0266} & 0.0283 & \textbf{0.0327} & \textbf{0.0571} & 0.0300 & \textbf{0.0155} \\
\addlinespace[2pt]
\multicolumn{7}{@{}l}{\textbf{Four-block harmonic}} \\
Fidelity $\mathcal M$ & 0.9752 & \textbf{0.9730} & \textbf{0.9662} & \textbf{0.9299} & \textbf{0.9714} & \textbf{0.9814} \\
Force error $e_F$ & \textbf{0.0231} & \textbf{0.0267} & \textbf{0.0362} & \textbf{0.0805} & \textbf{0.0285} & \textbf{0.0221} \\
Torque error $e_\tau$ & 0.0273 & \textbf{0.0282} & 0.0328 & 0.0674 & \textbf{0.0297} & \textbf{0.0155} \\
\bottomrule
\end{tabular*}
\par\vspace{6pt}
\begin{tabular*}{\textwidth}{@{\extracolsep{\fill}}lrrr@{}}
\toprule
\multicolumn{4}{@{}l}{\textbf{(b) Overall fidelity and computational cost}} \\
\textbf{Configuration} & \textbf{Mean fidelity} & \textbf{Speedup ($\times$)} & \textbf{Training (s)} \\
\midrule
Hover CA & \textbf{0.659313} & 67.51 & 22.5 \\
LQ wind-lag & \textbf{0.828039} & 68.29 & 137.3 \\
Mean-only & \textbf{0.898705} & 66.25 & 137.8 \\
Five-block harmonic & \textbf{0.964630} & 65.19 & 160.5 \\
Four-block harmonic & \textbf{0.966191} & 67.98 & 165.0 \\
\bottomrule
\end{tabular*}
\end{table*}

\topic{Where the Built-In Hover Model Succeeds and Fails.}
As shown in Table~\ref{tab:main_hierarchy_results} and Figure~\ref{fig:per_family_fidelity}(a), FlapSim's built-in cycle-averaged model reproduces the $50\,\mu\text{s}$ blade-element simulator well in still-air hover (frequency fidelity $0.9275$, $e_F = 0.0838$, $e_\tau = 0.0669$) and constant-tilt attitude control (gimbal fidelity $0.9440$, $e_F = 0.0508$, $e_\tau = 0.0647$).
However, its fidelity decreases in four regimes: (1) \textbf{3D Steady Wind (wind fidelity $0.2356$):} ignoring ambient airflow $\mathbf{w}$ yields $e_F = 1.8124$ (physical RMSE $37.83\,\text{mN}$) and $e_\tau = 1.3454$ ($1.966\,\text{mN}\cdot\text{m}$), approximately $1.81\times$ the reference force normalization scale;
(2) \textbf{Mid-Flight Frequency Steps (frequency-step fidelity $0.4874$):} instantaneous changes in $\bar{F}_z \propto f^2$ without clock synchronization or harmonic synthesis produce $e_F = 0.6922$ ($15.14\,\text{mN}$) and $e_\tau = 0.7563$ ($1.142\,\text{mN}\cdot\text{m}$);
(3) \textbf{Unsteady Wind Gusts (gust fidelity $0.6153$):} unmodeled $1-\cos$ gust impingement yields $e_F = 0.5810$ ($16.10\,\text{mN}$) and $e_\tau = 0.4399$ ($0.847\,\text{mN}\cdot\text{m}$);
and (4) \textbf{Parametric Wing Geometries (flat-plate fidelity $0.7462$):} extrapolating a single hover coefficient via $c R^3$ leaves a vertical thrust residual $e_F = 0.6030$ ($9.25\,\text{mN}$) despite low zero-gimbal torque error $e_\tau = 0.0184$ ($0.019\,\text{mN}\cdot\text{m}$) \citep{Fu_2018}.

\begin{figure*}[t]
\centering
\includegraphics[width=\textwidth]{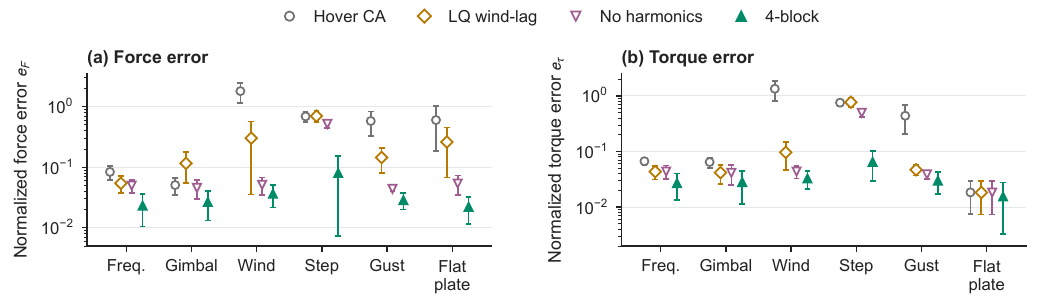}
\caption{Force and torque errors by flight regime. \textbf{(a)} Normalized force error $e_F$. \textbf{(b)} Normalized torque error $e_\tau$. Both axes are logarithmic. The original benchmark records a $50.1\times$ wind force-error reduction ($1.8124\to0.0362$) and a $27.3\times$ flat-plate force-error reduction ($0.6030\to0.0221$). Points and error bars show means and $\pm1$ sample standard deviation across the $8$ scenarios per regime in the local re-evaluation.}
\label{fig:force_torque_errors}
\end{figure*}

\topic{Closing the 3D Wind, Gust, and Planform Scaling Gaps.}
Figure~\ref{fig:force_torque_errors} decomposes normalized force error $e_F$ and torque error $e_\tau$ on a logarithmic scale across all six families. Introducing the compact first-principles linear/quadratic wind-drag and $12\,\text{ms}$ lag baseline lifts overall $48$-scenario fidelity by $+0.1687$ to $0.828039$ ($68.29\times$ speedup): wind fidelity increases from $0.2356$ to $0.8250$ ($e_F$ reduced $6.0\times$ from $1.8124$ to $0.3035$; $e_\tau$ reduced $13.9\times$ from $1.3454$ to $0.0967$), gust fidelity rises from $0.6153$ to $0.9088$, and flat-plate rises from $0.7462$ to $0.8734$.
Expanding to the full multi-block inflow and planform formulation reduces steady wind force error by $50.1\times$ relative to the hover baseline ($e_F = 1.8124 \to 0.0362$, physical RMSE $37.83\,\text{mN} \to 0.77\,\text{mN}$) and wind torque error by $41.1\times$ ($e_\tau = 1.3454 \to 0.0327$, physical RMSE $1.966\,\text{mN}\cdot\text{m} \to 0.048\,\text{mN}\cdot\text{m}$).
Similarly, multi-rate filtered gust states reduce gust force error by $20.4\times$ ($0.5810 \to 0.0285$, fidelity $0.9714$), and the polynomial planform basis reduces flat-plate force error by $27.3\times$ ($0.6030 \to 0.0221$, physical RMSE $9.25\,\text{mN} \to 0.30\,\text{mN}$, fidelity $0.9814$).

\subsection{Component Ablations: Harmonic Synthesis and Transient Lag States}
\begin{figure*}[t]
\centering
\includegraphics[width=\textwidth]{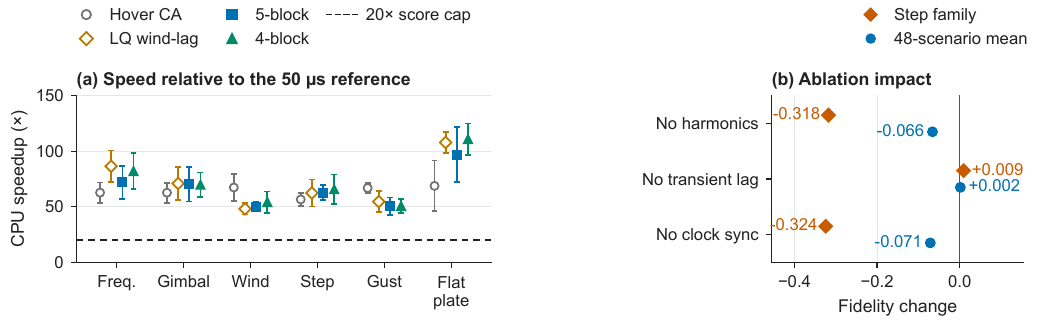}
\caption{Computational speedup and component ablations. \textbf{(a)} CPU speedup by regime, relative to the reference simulator; the dashed line marks the $20\times$ score-saturation threshold. Points and error bars show means and $\pm1$ sample standard deviation across the $8$ scenarios per regime in the local re-evaluation. Original benchmark family speedups span $52.07\times$ to $109.85\times$. \textbf{(b)} Changes in frequency-step and overall fidelity when removing harmonics, explicit transient states, or clock synchronization. Frequency-step changes are $-0.3176$, $+0.0095$, and $-0.3243$, respectively. Panel (b) shows aggregate benchmark differences and has no error bars; the clock comparison uses the completed pre- and post-synchronization runs described in the text.}
\label{fig:speedup_and_ablations}
\end{figure*}

\begin{figure*}[t]
\centering
\includegraphics[width=\textwidth]{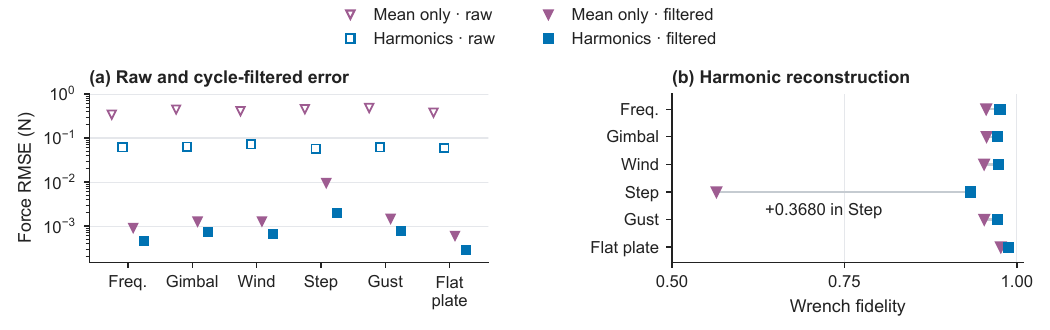}
\caption{Harmonic reconstruction and moving-average filter leakage. \textbf{(a)} Raw and cycle-filtered three-dimensional force RMSE. Harmonic synthesis reduces raw RMSE by $6.65\times$ ($0.4166\,\text{N}\to0.0627\,\text{N}$). \textbf{(b)} Mean-only and harmonic-model fidelity by regime in the original post-hoc evaluation. The frequency-step improvement is $+0.3680$, consistent with cancellation of the $244.2\times$ increase in fundamental-frequency transmission at $f_{\text{high}}$. Values are aggregate results; neither panel displays standard-deviation bars.}
\label{fig:posthoc_02_harmonics}
\end{figure*}

We next examine why a cycle-averaged evaluation benchmark---which explicitly filters both reference and surrogate wrenches with a two-pass one-stroke-period moving average $\mathcal{F}_W$---is sensitive to within-cycle stroke harmonics during frequency steps. Figure~\ref{fig:speedup_and_ablations} and Figure~\ref{fig:posthoc_02_harmonics} present the controlled ablation results and mathematical transfer-function analysis.

\topic{1. Ablating $1f\text{--}6f$ Stroke-Phase Harmonics.}
Zeroing all within-cycle harmonic weights ($\mathbf{W}_h^{(b)} = \mathbf{0}$) while retaining all cycle-averaged mean blocks, $21$-substep clock synchronization, and Block-3 step-transient mean filters reduces the $48$-scenario composite score by $\Delta S = -0.065925$ (from $0.964630$ to $0.898705$).
Across the five constant-frequency families, removing harmonics causes only a modest fidelity reduction of $0.0118\text{--}0.0212$ (frequency $0.9755 \to 0.9543$, gimbal $0.9730 \to 0.9576$, wind $0.9662 \to 0.9535$, gust $0.9712 \to 0.9594$, flat-plate $0.9814 \to 0.9646$). In comparison, during mid-flight frequency steps, removing harmonics changes fidelity by $-0.3176$ on the held-out test set ($0.9204 \to 0.6028$; force error increasing $4.76\times$ from $0.1095$ to $0.5209$, torque error increasing $8.69\times$ from $0.0571$ to $0.4960$) and by $-0.3680$ on the separate $48$-scenario post-hoc evaluation suite ($0.9325 \to 0.5645$, paired $t = 10.07$, $p = 2.04 \times 10^{-5}$, exact Wilcoxon $p = 0.0078125$).

\topic{2. Transfer-Function Analysis of Moving-Average Filter Leakage.}
For a harmonic oscillating at $f_{\text{osc}} = k f$, two passes of a discrete boxcar moving average of integer length $W$ have the exact amplitude transfer gain:
\begin{equation}
    G_{\text{2pass}}(k f, W) = \left(\frac{\sin(\pi k f W \Delta t)}{W \sin(\pi k f \Delta t)}\right)^{\!2}.
    \label{eq:filter_gain}
\end{equation}
Post-hoc evaluation across all $48$ scenarios (Figure~\ref{fig:posthoc_02_harmonics}) reveals two physical regimes: (i) \textbf{Constant-Frequency Integer-Millisecond Rounding Leakage ($n=40$ scenarios):} because $W = \text{round}(1000/f)$ rounds the stroke period to the nearest integer millisecond, $W f \Delta t = 1 + \epsilon$ with $|\epsilon| \le 1/(2W) \approx 1.1\%\text{--}2.4\%$, yielding a non-zero two-pass transmission $G_{\text{2pass}}(1f, W) = 1.25 \times 10^{-4}$ on average. Because the raw blade-element flapping force oscillation has an RMS amplitude of $0.34\text{--}0.48\,\text{N}$ ($15\times\text{--}21\times$ larger than $F_{\text{scale}} \approx 0.0225\,\text{N}$), the evaluation records $0.6\text{--}1.5\,\text{mN}$ of filtered error in pure mean models. This aggregate residual also depends on higher harmonics and time-varying loads; it cannot be inferred from the fundamental-frequency gain alone. Synthesizing phase-locked $1f\text{--}6f$ harmonics passes the matching waveform through $\mathcal{F}_W$, cutting filtered force RMSE in half ($1.73\times\text{--}2.15\times$ reduction) and yielding a $+0.0173$ mean fidelity lift across the $40$ constant-frequency scenarios ($95\%$ CI $[+0.0140, +0.0208]$, paired $t = 9.70$, $p = 6.06 \times 10^{-12}$).
(ii) \textbf{Mid-Flight Frequency-Step Period Mismatch and Phase-Jump Leakage ($n=8$ scenarios):} in frequency-step scenarios, the stroke frequency jumps at $t = 0.3\,\text{s}$ between $f_1$ and $f_2$, but the evaluator applies a \emph{single fixed window} $W = \text{round}(1000 / \min(f_1, f_2))$ tuned exclusively to $f_{\text{low}} = \min(f_1, f_2)$. During the $300\,\text{ms}$ interval where the wing flaps at $f_{\text{high}} = \max(f_1, f_2)$, $W$ spans $W f_{\text{high}} \Delta t \in [1.14, 1.70]$ stroke cycles. Consequently, single-pass $1f_{\text{high}}$ transmission increases to $11.19\%\text{--}21.09\%$ (mean $15.59\%$) and two-pass transmission $G_{\text{2pass}}(1f_{\text{high}}, W)$ increases from $1.03 \times 10^{-4}$ to $0.02515$---a \textbf{$244.2\times$ increase} in $1f$ transmission and a \textbf{$105.9\times$ increase} in $2f$ transmission.
Combined with the $2W$-sample window straddling the phase jump at $t = 0.3\,\text{s}$, this leaks $9.56\,\text{mN}$ of harmonic oscillation ($\approx 43\%$ of $F_{\text{scale}}$) into the filtered reference trajectory. Phase-locked $1f\text{--}6f$ synthesis reduces filtered frequency-step force RMSE by $4.88\times$ ($9.56\,\text{mN} \to 1.96\,\text{mN}$) while simultaneously reconstructing the unfiltered $50\,\mu\text{s}$ within-stroke waveform across all $48$ scenarios, reducing raw unfiltered 3D force RMSE by $6.65\times$ ($0.4166\,\text{N} \to 0.0627\,\text{N}$, paired $t = 15.05$, $p < 10^{-15}$) and raw torque RMSE by $3.35\times$ ($0.0210\,\text{N}\cdot\text{m} \to 0.00625\,\text{N}\cdot\text{m}$).

\topic{3. Redundancy of Explicit Step-Transient Lag States and Multirate Clock Alignment.}
Whereas removing stroke harmonics reduces frequency-step fidelity, removing the $104$-feature Block-3 frequency-step transient module (setting $\mathbf{W}_{m}^{(3)} = \mathbf{0}$ and $\mathbf{W}_{h}^{(3)} = \mathbf{0}$) slightly \emph{improves} held-out $48$-scenario mean fidelity from $0.964630$ to $0.966191$ ($\Delta S = +0.001561$), lifting held-out frequency-step fidelity from $0.9204$ to $0.9299$ (reducing frequency-step force error $e_F$ from $0.1095$ to $0.0805$) and increasing CPU speedup from $65.19\times$ to $67.98\times$.
This result reveals that once steady-form $1f\text{--}6f$ harmonics (Blocks 1 \& 2) and $21$-substep clock synchronization are active, quasi-static switching between the calibrated pre-step and post-step harmonic basis functions already captures the moving-average filter transition accurately in this benchmark across $t = 0.3\,\text{s}$, whereas the slight degradation with Block 3 is consistent with overfitting of training step magnitudes. Conversely, omitting the $21$-substep clock synchronization law (Eq.~\ref{eq:clock_sync}) leaves a $5\%$ time-base drift that reduces overall fidelity from $0.964536$ to $0.893374$ ($\Delta S = -0.0712$) and drops frequency-step fidelity to $0.5957$ (Figure~\ref{fig:speedup_and_ablations}(b)).

\subsection{Per-Scenario Wrench Error Clustering, Asymmetry, and Limitations}
\begin{figure*}[t]
\centering
\includegraphics[width=\textwidth]{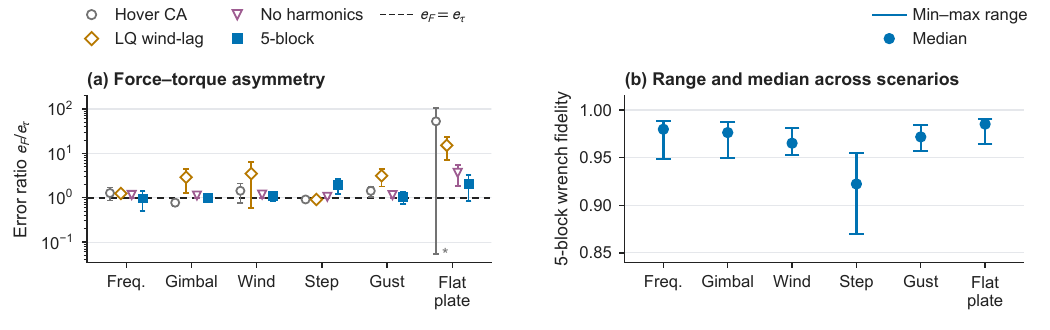}
\caption{Force--torque asymmetry and fidelity dispersion. \textbf{(a)} Mean scenario-wise error ratio $e_F/e_\tau$ with $\pm1$ sample standard deviation ($n=8$ per regime). \textbf{(b)} Minimum--maximum ranges (lines) and medians (points) of the five-block model's fidelity across those scenarios. Both panels use the local re-evaluation. The original post-hoc evaluation discussed in the text is a separate record: $100\%$ of scenarios below its bottom-decile threshold $P_{10}=0.9435$ are frequency-step cases, with a minimum of $0.8924$ at scenario \#28. The starred lower SD bar for the hover baseline in the flat-plate regime crosses zero and is displayed at $0.001$ times its mean on the logarithmic axis; its upper SD is unchanged.}
\label{fig:posthoc_01_asymmetry}
\end{figure*}

Beyond family averages, Figure~\ref{fig:posthoc_01_asymmetry} summarizes our scenario-level error clustering and force-versus-torque asymmetry analysis across all $48$ evaluation trajectories.

\topic{1. Concentration of Tail Risk and Force-vs-Torque Asymmetry.}
In the original $48$-scenario post-hoc evaluation (a separate seed set from the held-out benchmark), $100\%$ of the bottom-decile tail risk ($5$ out of $5$ lowest-fidelity scenarios below the 10th percentile cutoff $P_{10} = 0.9435$) in the harmonic surrogate is concentrated in the frequency-step family (scenarios \#28, \#25, \#31, \#32, and \#29).
The single worst-case scenario is frequency-step \#28 (seed $22193$, stepping upward from $f_1 = 25.04\,\text{Hz}$ to $f_2 = 44.45\,\text{Hz}$ with filter window duration $W\Delta t = 40\,\text{ms}$), which records a fidelity of $0.8924$ with strongly force-dominated residual asymmetry ($e_F = 0.1507$ vs. $e_\tau = 0.0771$, ratio $e_F / e_\tau = 1.954$; held-out test frequency-step ratio $e_F / e_\tau = 1.918$ in the five-block harmonic model and $1.194$ in the four-block harmonic model).
In the flat-plate family, error is structurally force-dominated across all surrogates (held-out $e_F / e_\tau = 1.43$ in the harmonic surrogates vs. $14.26$ in the first-principles linear/quadratic baseline and $32.77$ in the hover baseline) because symmetric vertical thrust errors at zero gimbal tilt do not induce roll or pitch moments.
In comparison, across the $24$ 3D inflow and gimbal scenarios (gimbal, wind, gust), our harmonic surrogate restores tightly balanced 6-DoF residuals (held-out $e_F / e_\tau \in [0.94, 1.11]$, with Pearson correlation $r = 0.9768$ between physical force and torque RMSEs and $r = 0.8568$ between physical force RMSE and the reference force scale $F_{\text{scale}}$).

\topic{2. Discussion of Platform Limitations and Roadmap.}
While the streamlined four-block harmonic surrogate achieves $0.9662$ mean fidelity across all $48$ scenarios at a $67.98\times$ CPU speedup, three limitations define the scope of our findings and motivate future platform development: (i) \textbf{Fixed-Base Protocol vs. Free-Flight Coupling:} all $48$ benchmark scenarios evaluate aerodynamic wrenches on a fixed base with prescribed wing strokes following wind-tunnel practice \citep{Wang_2016, Pohly_2018}; while local body velocities enter Eq.~\eqref{eq:strip_velocity}, closed-loop free-flight maneuvers \citep{Cheng_2016} introduce attitude feedback that should be benchmarked via FlapSim's Gymnasium simulation environment wrapper; (ii) \textbf{Multirate Substep Clock Semantics:} our identification of the $21$-substep ($1.05\,\text{ms}$) reference clock advancement for $t \ge 0.2\,\text{s}$ highlights a floating-point pitfall when driving a $50\,\mu\text{s}$ reference simulator inside a $1\,\text{ms}$ recording loop, which motivates integer substepping ($20$ substeps per recording interval) and a separate re-evaluation under the corrected clock; and (iii) \textbf{Calibration Complexity vs. Generalization:} higher-complexity calibration schemes---filter-projected scale-normalized IRLS/L-BFGS-B refinement (scoring $0.963339$ at $67.49\times$ speedup) and exact rank-1 leave-one-scenario-out (LOSO) hybrid passband-stopband calibration (scoring $0.963467$ at $65.53\times$ speedup)---did not outperform the simpler ridge-calibrated four-block harmonic hierarchy ($0.966191$).
Of $11$ total model configurations evaluated during the study, $10$ completed validly and $1$ terminated with a score of $0.0$ due to a Python multiprocessing process-spawning module-name mismatch prior to switching to forked workers; similarly, in the six-variant ablation runner, $5$ variants completed full evaluation and $1$ standalone nominal-clock ablation variant had zero completed runs (completed evaluations $=0$), so the clock-synchronization delta is grounded in the completed pre-synchronization versus post-synchronization 5-block evaluations ($0.893374$ vs. $0.964536$).

\section{Conclusion}
FlapSim v0.2.0 combines a modular MuJoCo 3.8 platform with an aerodynamic model hierarchy evaluated across $48$ fixed-base scenarios in six regimes. Against the $50\,\mu\mathrm{s}$ blade-element reference, the $1\,\mathrm{ms}$ hover baseline achieves $0.6593$ mean wrench fidelity at a $67.51\times$ CPU speedup. Combining clock alignment, inflow and gust coupling, planform scaling, and phase-locked harmonics raises fidelity to $0.9646$ ($65.19\times$ speedup) in the five-block model and $0.9662$ ($67.98\times$) in the four-block model without explicit step-transient states.

Harmonic synthesis reduces raw force RMSE by approximately $85\%$ relative to the mean-only model ($6.65\times$ reduction) and reproduces oscillations retained by the fixed-window filter during frequency steps, where fundamental-frequency transmission increases by $244.2\times$. These results establish the accuracy--computation tradeoff within the evaluated protocol. Generalization to a corrected reference clock, closed-loop flight, and reinforcement-learning tasks remains to be evaluated.

\section*{LLM Usage Statement}
Coding agents and large language models assisted with plugin implementation, evaluation in the fixed FlapSim harness, reference retrieval, post-hoc statistical analysis, telemetry-based figures, and LaTeX drafting and validation. Reported metrics, codebase counts, and statistical tests are grounded in recorded simulator outputs in \id{paper_data.json}.

\bibliography{references}
\bibliographystyle{IEEEtran}

\par\bigskip
\appendices
\section{Supplementary Information}

\subsection{Complete FlapSim v0.2.0 Codebase, Plugin, and Robot Manifest Inventory}
Table~\ref{tab:si_plugins_manifests} lists all $17$ registered plugins across the seven plugin kinds of FlapSim v0.2.0, together with the $4$ aerodynamic models and $1$ modifier available to the composite aerodynamic model, the robot descriptions shipped with the package, and the verification suite ($117$ test functions across $23$ test files).

\begin{table*}[!t]
\caption{Complete registry of FlapSim v0.2.0 plugins ($17$ across $7$ kinds), aerodynamic models ($4$) and modifiers ($1$), shipped robot descriptions, and verification suite.}
\label{tab:si_plugins_manifests}
\centering
\small
\setlength{\tabcolsep}{4pt}
\renewcommand{\arraystretch}{1.13}
\begin{tabularx}{\textwidth}{@{}p{.17\textwidth}cY@{}}
\toprule
\textbf{Kind / category} & \textbf{Count} & \textbf{Registered identifiers and scope} \\
\midrule
\id{actuator} & 4 & \id{sinusoidal_stroke} (kinematic or position-only stroke), \id{pd_position} (PD or kinematic hold), \id{servo} (rate-limited lagged target), \id{dc_motor_crank} (voltage-driven motor, gearbox, crank) \\
\addlinespace[2pt]
\id{aero} & 3 & \id{quasi_steady} (single-model stack), \id{stack} (additive models and modifiers), \id{cycle_averaged} (rigid-body hover model, identified parameters) \\
\addlinespace[2pt]
\id{atmosphere} & 1 & \id{standard} (ISA density or $\rho$, steady wind, $1-\cos$/sine/step gust, Ornstein--Uhlenbeck turbulence) \\
\addlinespace[2pt]
\id{controller} & 3 & \id{hover_pid} (cascaded altitude and attitude PID), \id{schedule} (time-scheduled set points), \id{callback} (user Python function) \\
\addlinespace[2pt]
\id{force} & 1 & \id{body_drag} (quadratic drag per body axis relative to the wind) \\
\addlinespace[2pt]
\id{recorder} & 1 & \id{memory} (samples matching signals every $\Delta t$; export) \\
\addlinespace[2pt]
\id{sensor} & 4 & \id{imu}, \id{rangefinder}, \id{encoder}, \id{wrench}, each with sample-and-hold, low-pass, delay, noise, bias and quantization conditioning \\
\addlinespace[2pt]
\midrule
Aerodynamic models & 4 & \id{quasi_steady} (Whitney \& Wood 2010; Dickinson et al. fits), \id{rotational_lift}, \id{added_mass}, \id{wake_capture} \\
\addlinespace[2pt]
Aerodynamic modifiers & 1 & \id{ground_effect} (exponential or Cheeseman--Bennett factor) \\
\addlinespace[2pt]
Robot descriptions & 1 + builder & \id{SingleThrustFlapper} ($30\,\text{g}$, $R = 70\,\text{mm}$; URDF, \id{robot.yaml}, \id{sim.yaml}, \id{hover.yaml}) and the parametric \id{flat_plate_flapper} builder \\
\addlinespace[2pt]
Verification suite & 117 & $117$ unit, regression (lift sweep), conformance and example tests across $23$ files; $22$ documentation pages; $24$ example scripts \\
\bottomrule
\end{tabularx}
\end{table*}

\subsection{Physical Force (N) and Torque (N m) RMSE and Runtime Breakdown by Scenario Family}
While Table~\ref{tab:main_hierarchy_results} in the main text reports normalized force error $e_F$ and normalized torque error $e_\tau$, Table~\ref{tab:si_physical_rmse} provides the complete dimensional force RMSE ($\text{mN}$), dimensional torque RMSE ($\text{mN}\cdot\text{m}$), per-family CPU speedup ($\times$), total candidate and reference CPU seconds, step throughput (steps/s), and offline training and testing wall-clock times for all five completed ablation-study variants in \id{paper_data.json}.

\begin{table*}[!t]
\caption{Cycle-filtered errors and runtime over $48$ held-out scenarios; 5-block and 4-block denote harmonic models. Force/torque RMSE units are $\mathrm{mN}$/$\mathrm{mN}\!\cdot\!\mathrm m$ ($10^{-3}\,\mathrm N$/$10^{-3}\,\mathrm N\!\cdot\!\mathrm m$). CPU speedup uses the blade-element reference.}
\label{tab:si_physical_rmse}
\centering
\small
\setlength{\tabcolsep}{4pt}
\renewcommand{\arraystretch}{1.13}
\renewcommand{\arraystretch}{1.05}
\begin{tabular*}{\textwidth}{@{\extracolsep{\fill}}llrrrrrr@{}}
\toprule
\multicolumn{8}{@{}l}{\textbf{(a) Dimensional errors and per-family speedup}} \\
\textbf{Model} & \textbf{Metric} & \textbf{Frequency} & \textbf{Gimbal} & \textbf{Wind} & \textbf{Step} & \textbf{Gust} & \textbf{Flat plate} \\
\midrule
\multirow{3}{*}{\textbf{Hover CA}} & Force RMSE ($\text{mN}$) & 1.50 & 1.52 & 37.83 & 15.14 & 16.10 & 9.25 \\
 & Torque RMSE ($\text{mN}\cdot\text{m}$) & 0.086 & 0.133 & 1.966 & 1.142 & 0.847 & 0.019 \\
 & Family Speedup ($\times$) & 65.83 & 82.09 & 65.38 & 61.97 & 60.40 & 87.20 \\
\addlinespace[2pt]
\multirow{3}{*}{\textbf{LQ wind-lag}} & Force RMSE ($\text{mN}$) & 1.04 & 3.57 & 6.09 & 15.20 & 3.92 & 3.72 \\
 & Torque RMSE ($\text{mN}\cdot\text{m}$) & 0.057 & 0.087 & 0.145 & 1.153 & 0.094 & 0.019 \\
 & Family Speedup ($\times$) & 81.27 & 69.96 & 53.26 & 66.90 & 63.17 & 97.29 \\
\addlinespace[2pt]
\multirow{3}{*}{\textbf{Mean-only}} & Force RMSE ($\text{mN}$) & 0.93 & 1.36 & 1.07 & 11.18 & 1.26 & 0.76 \\
 & Torque RMSE ($\text{mN}\cdot\text{m}$) & 0.057 & 0.086 & 0.063 & 0.746 & 0.078 & 0.019 \\
 & Family Speedup ($\times$) & 77.72 & 75.78 & 55.55 & 56.49 & 59.32 & 115.69 \\
\addlinespace[2pt]
\multirow{3}{*}{\textbf{5-block}} & Force RMSE ($\text{mN}$) & 0.47 & 0.80 & 0.77 & 2.27 & 0.84 & 0.30 \\
 & Torque RMSE ($\text{mN}\cdot\text{m}$) & 0.036 & 0.059 & 0.048 & 0.084 & 0.060 & 0.015 \\
 & Family Speedup ($\times$) & 86.57 & 77.33 & 52.69 & 64.31 & 52.07 & 92.68 \\
\addlinespace[2pt]
\multirow{3}{*}{\textbf{4-block}} & Force RMSE ($\text{mN}$) & 0.47 & 0.80 & 0.77 & 1.70 & 0.84 & 0.30 \\
 & Torque RMSE ($\text{mN}\cdot\text{m}$) & 0.037 & 0.058 & 0.048 & 0.099 & 0.060 & 0.015 \\
 & Family Speedup ($\times$) & 79.20 & 66.03 & 52.57 & 68.92 & 61.38 & 109.85 \\
\bottomrule
\end{tabular*}
\par\vspace{4pt}
\begin{tabular*}{\textwidth}{@{\extracolsep{\fill}}lrrrrrrr@{}}
\toprule
\multicolumn{8}{@{}l}{\textbf{(b) Overall runtime summary}} \\
& \multicolumn{2}{c}{\textbf{Candidate}} & \multicolumn{2}{c}{\textbf{Reference}} & \multicolumn{2}{c}{\textbf{Wall time}} & \\
\cmidrule(lr){2-3}\cmidrule(lr){4-5}\cmidrule(lr){6-7}
\textbf{Configuration} & CPU (s) & Steps/s & CPU (s) & Steps/s & Train (s) & Test (s) & Speedup ($\times$) \\
\midrule
Hover CA & 2.173 & 13{,}027.6 & 146.68 & 3{,}987.4 & 22.5 & 20.9 & 67.506 \\
LQ wind-lag & 2.054 & 13{,}780.0 & 140.28 & 4{,}169.2 & 137.3 & 18.4 & 68.290 \\
Mean-only & 2.115 & 13{,}384.5 & 140.12 & 4{,}174.1 & 137.8 & 18.8 & 66.252 \\
5-block & 2.031 & 13{,}936.8 & 132.40 & 4{,}417.2 & 160.5 & 17.9 & 65.189 \\
4-block & 2.055 & 13{,}774.8 & 139.70 & 4{,}186.5 & 165.0 & 18.7 & 67.982 \\
\bottomrule
\end{tabular*}
\end{table*}

\subsection{Mathematical Analysis of Multirate 21-Substep Clock Drift and Filter-Window Non-Commutativity}
\topic{1. Derivation of the 21-Substep Reference Clock Drift.}
At the start of recording call $k$, let $t_k$ be the accumulated simulation time. The implemented advance routine sets the relative target $t_{\mathrm{target},k}=t_k+10^{-3}\,\mathrm{s}$ and repeatedly increments time by $\Delta t_{\mathrm{ref}}=5\times10^{-5}\,\mathrm{s}$ until this target is reached. Floating-point rounding can require $21$ rather than $20$ substeps. Since the next target is based on the newly accumulated time, these overruns accumulate; a target anchored to $0.2+(k+1)10^{-3}$ would instead prevent this cumulative dilation. Approximating the repeated calls by $21$ substeps gives $1.05\,\mathrm{ms}$ per recording call and about $30\,\mathrm{ms}$ of drift over $600$ calls ($0.75$--$1.35$ wingbeat cycles at $25$--$45\,\mathrm{Hz}$). Exact substep counts depend on the floating-point trajectory, so Eq.~\eqref{eq:clock_sync} is an implementation-specific alignment approximation, not a general identity of IEEE-754 arithmetic. Frequency changes are scheduled by recording index; their reference-clock time consequently differs from the nominal timestamp. The completed pre- and post-alignment runs report frequency-step fidelity of $0.5957$ and $0.9204$, respectively, but the isolated nominal-clock ablation did not complete.

\topic{2. Non-Commutativity of Fixed-Window Moving-Average Filtering Across Frequency Steps.}
Let $W = \text{round}(1 / (f_{\text{low}}\Delta t))$ denote the fixed moving-average window length for a frequency-step scenario with pre-step frequency $f_1$ and post-step frequency $f_2$, where $f_{\text{low}} = \min(f_1, f_2)$ and $f_{\text{high}} = \max(f_1, f_2)$. Consider the $k$-th harmonic component of the blade-element wrench, $h_k(t) = A_k(t)\cos(2\pi k f(t) t + \varphi_k)$, where $f(t) = f_1$ for $t < t_{\text{step}}$ and $f(t) = f_2$ for $t \ge t_{\text{step}}$. Applying the two-pass discrete moving-average operator $\mathcal{F}_W$ yields three distinct regimes:
\begin{enumerate}
    \item \textbf{Low-frequency segment ($f(t) = f_{\text{low}}$, $|t - t_{\text{step}}| > W\Delta t$):} Because $W f_{\text{low}}\Delta t \approx 1$, the recorded mean fundamental gain is $G_{\text{2pass}}(f_{\text{low}}, W) \approx 1.03 \times 10^{-4}$, strongly attenuating the harmonic.
    \item \textbf{Straddle window ($|t - t_{\text{step}}| \le W\Delta t$):} The averaging window sums samples from both $f_1$ and $f_2$ across the phase discontinuity $\Delta\psi = 2\pi(f_2 - f_1)t_{\text{step}}$, producing a non-zero transient pulse of duration $2W\Delta t \in [52, 80]\,\text{ms}$ whose sign and magnitude depend on the exact phase at $t_{\text{step}}$.
    \item \textbf{High-frequency segment ($f(t) = f_{\text{high}}$, $|t - t_{\text{step}}| > W\Delta t$):} Because $W f_{\text{high}}\Delta t \approx f_{\text{high}} / f_{\text{low}} \in [1.14, 1.70]$ is generally non-integer, the filter need not suppress the harmonics, with the recorded mean increasing two-pass transmission by $244.2\times$ at $1f_{\text{high}}$ (mean $G_{\text{2pass}} = 0.02515$) and $105.9\times$ at $2f_{\text{high}}$.
\end{enumerate}
Thus, $\mathcal{F}_W[h_k](t)$ can be nonzero in both $|t - t_{\text{step}}| \le W\Delta t$ and the entire $f_{\text{high}}$ half of the trajectory. Because $\mathcal{F}_W$ is a linear operator, generating the matching phase-locked harmonics $\tilde{\mathbf{W}}(t)$ inside the $1\,\text{ms}$ surrogate guarantees $\mathcal{F}_W[\bar{\mathbf{W}} + \tilde{\mathbf{W}}] = \mathcal{F}_W[\bar{\mathbf{W}}] + \mathcal{F}_W[\tilde{\mathbf{W}}]$, so accurate harmonic reconstruction can reproduce both the straddle pulse and the leaked $f_{\text{high}}$ oscillation; linearity alone does not guarantee that the fitted harmonics are accurate.

\subsection{Multi-Block Ridge Weight Norms, Cross-Validation Sensitivity, and Training Budget Analysis}
\begin{figure*}[t]
\centering
\includegraphics[width=\textwidth]{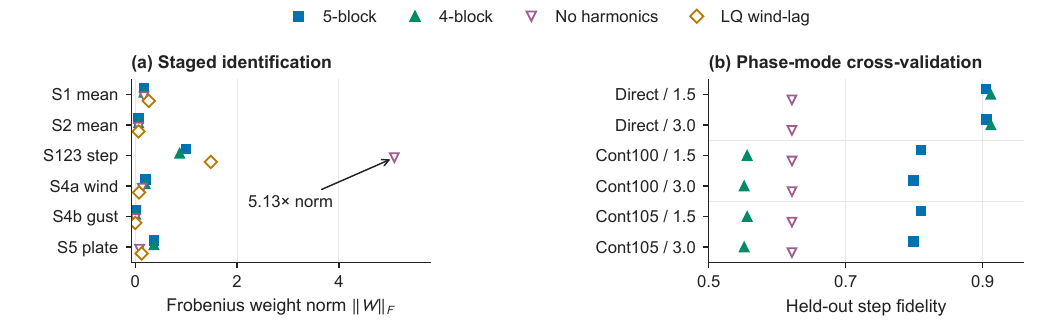}
\caption{Ridge-regression weight norms and sensitivity to the phase model. \textbf{(a)} Frobenius weight norm $\|\mathbf W\|_F$ by identification stage. Subtracting the Block 1--2 harmonic predictions before Step 5 avoids a $5.13\times$ increase in weight norm ($0.9920$ for the five-block model versus $5.0885$ without harmonics). \textbf{(b)} Held-out frequency-step cross-validation fidelity for six combinations of phase convention (direct phase, or continuous phase at nominal or synchronized rate) and harmonic sample weight ($1.5$ or $3.0$). Points are the recorded fit results; no standard-deviation bars are shown.}
\label{fig:posthoc_03_training}
\end{figure*}

Figure~\ref{fig:posthoc_03_training} and Table~\ref{tab:si_staged_regression} present the complete offline identification and cross-validation analysis.

\begin{table*}[!t]
\caption{Offline ridge identification: fit wall time (s), Frobenius norm $\|\mathbf W\|_F$, and normalized training RMSE. 5B/4B: five-/four-block harmonic; MO: mean-only. RMSE ratio: MO/5B.}
\label{tab:si_staged_regression}
\centering
\footnotesize
\setlength{\tabcolsep}{4pt}
\renewcommand{\arraystretch}{1.13}
\setlength{\tabcolsep}{2.4pt}
\renewcommand{\arraystretch}{1.05}
\begin{minipage}[t]{.44\textwidth}
\vspace{0pt}
\textbf{(a) Identification stages and fit times}\par\vspace{2pt}
\begin{tabularx}{\linewidth}{@{}cYr@{}}
\toprule
\textbf{Step} & \textbf{Block and active features} & \textbf{Fit (s)} \\
\midrule
1 & Pass-1 Block 1 Hover Mean (\id{w_m1}, 12 feat.) & 0.048 \\
\addlinespace[1pt]
2 & Pass-1 Block 2 Gimbal Mean (\id{w_m2}, 27 feat.) & 0.413 \\
\addlinespace[1pt]
3 & Pass-2 Joint Gauss--Seidel \id{w_m1} (12 feat.) & 0.361 \\
\addlinespace[1pt]
4 & Pass-2 Joint Gauss--Seidel \id{w_m2} (27 feat.) & 0.364 \\
\addlinespace[1pt]
5 & 6-Grid CV + Harmonics (\id{w_h1,2}, 66) + Block 3 (104) & 37.289 \\
\addlinespace[1pt]
6 & Block 4a Steady 3D Wind (54 mean, 24 harm.) & 1.439 \\
\addlinespace[1pt]
7 & Block 4b Unsteady 3D Gust (36 mean, 24 harm.) & 0.877 \\
\addlinespace[1pt]
8 & Block 5 Parametric Flat Plate (26 mean, 32 harm.) & 0.722 \\
\addlinespace[1pt]
\bottomrule
\end{tabularx}
\end{minipage}\hfill
\begin{minipage}[t]{.54\textwidth}
\vspace{0pt}
\textbf{(b) Weight norms and residuals}\par\vspace{2pt}
\begin{tabular*}{\linewidth}{@{\extracolsep{\fill}}crrrrrr@{}}
\toprule
& \multicolumn{3}{c}{\textbf{Weight norm}} & \multicolumn{2}{c}{\textbf{RMSE}} & \\
\cmidrule(lr){2-4}\cmidrule(lr){5-6}
Step & 5B & 4B & MO & 5B & MO & Ratio \\
\midrule
1 & 0.1639 & 0.1639 & 0.1639 & 0.01961 & 0.01961 & -- \\
2 & 0.0638 & 0.0638 & 0.0638 & 0.01920 & 0.01920 & -- \\
3 & 0.1708 & 0.1708 & 0.1708 & 0.01924 & 0.01924 & -- \\
4 & 0.0638 & 0.0638 & 0.0638 & 0.01919 & 0.01919 & -- \\
\textbf{5} & \textbf{0.9920} & \textbf{0.8718} & \textbf{5.0885} & \textbf{0.03621} & \textbf{0.17224} & \textbf{4.76} \\
6 & 0.1986 & 0.1943 & 0.1360 & 0.01334 & 0.02211 & 1.66 \\
7 & 0.0076 & 0.0055 & 0.0017 & 0.01328 & 0.02394 & 1.80 \\
8 & 0.3673 & 0.3673 & 0.0845 & 0.00649 & 0.01264 & 1.95 \\
\bottomrule
\end{tabular*}
\par\vspace{3pt}\raggedright Step 5: MO/5B weight-norm ratio $5.13\times$; residual RMSE ratio $4.76\times$.
\end{minipage}
\end{table*}

\topic{1. Offline Compute Partition and 8-Worker Parallelism.}
In the five-block harmonic model ($160.50\,\text{s}$ total training wall-clock time), $72.2\%$ ($115.96\,\text{s}$) is spent simulating the $356$ space-filling $50\,\mu\text{s}$ reference trajectories ($304$ gimbaled-flapper trajectories requiring $822.18\,\text{s}$ reference CPU time at $2.705\,\text{s}$ mean CPU time per trajectory, and $52$ flat-plate flapper trajectories requiring $104.36\,\text{s}$ reference CPU time at $2.007\,\text{s}$ per trajectory). Executing these $926.55\,\text{s}$ of cumulative reference CPU work across $8$ forked worker processes yields a $7.99\times$ parallel throughput gain ($7.09\times$ relative to the gimbaled flapper's CPU time alone), while the 8-step ridge regression pipeline consumes $41.51\,\text{s}$ ($25.9\%$, of which $37.29\,\text{s}$ is spent in Step 5's 6-candidate cross-validation grid) and post-fit verification across $48$ scenarios takes $3.03\,\text{s}$ ($1.9\%$).

\topic{2. Why Removing Harmonics Causes a $5.13\times$ Weight-Norm Increase in Step 5.}
As shown in Table~\ref{tab:si_staged_regression} and Figure~\ref{fig:posthoc_03_training}(a), subtracting the filtered Stage 1--2 harmonic predictions prior to fitting Block 3 in the five-block harmonic model keeps the Step-5 Frobenius weight norm small ($\|\mathbf{W}\|_F = 0.9920$, or $0.8718$ in the 4-block model) with a low normalized residual RMSE of $0.03621$. When harmonics are zeroed, the unsubtracted oscillatory filter leakage remains inside the Step-5 target residual, forcing Block 3's $48$ collinear real exponential-decay and low-pass-filter features ($\tau \in \{2, 5, 12, 25, 50, 100\}\,\text{ms}$) to fit oscillatory ringdown with large opposing weights. This increases $\|\mathbf{W}\|_F$ by $5.13\times$ to $5.0885$, leaves a $4.76\times$ larger training residual ($0.17224$), and coincides with a generalization gap (frequency-step fidelity dropping from $0.6953$ in-sample to $0.6215$ on CV validation and $0.6028$ on the held-out test set).

\topic{3. Cross-Validation Phase-Mode Selection.}
Across the 6-candidate CV grid (Figure~\ref{fig:posthoc_03_training}(b)), selecting direct phase ($\psi_d(t) = 2\pi f(t)t$ with harmonic sample weight $3.0$) achieves a held-out CV frequency-step fidelity of $0.90580$ in the five-block harmonic model and $0.91203$ in the four-block harmonic model. Without Block 3, switching to continuous phase at either the nominal or synchronized rate shifts post-step harmonics out of phase with the sinusoidal actuator, changing CV frequency-step fidelity by $-0.3603$ to $0.5518$ (span $0.3603$). With both phase conventions in Block 3, the five-block model spans $0.7989\text{--}0.9058$ CV fidelity (range $0.1069$, $3.37\times$ narrower).

\end{document}